\documentclass[10pt, a4paper]{article}
\usepackage{lrec}
\usepackage{multibib}
\newcites{languageresource}{Language Resources}
\usepackage{graphicx}
\usepackage{tabularx}
\usepackage{soul}

\usepackage{epstopdf}
\usepackage[utf8]{inputenc}

\usepackage{hyperref}
\usepackage{xstring}

\usepackage{url}
\usepackage{multirow}
\usepackage{graphicx}
\usepackage{comment}

\title{A Corpus for Multilingual Document Classification in Eight Languages}

\name{Holger Schwenk, Xian Li}
\address{Facebook AI Research, Facebook AML \\
	\{schwenk,xianl\}@fb.com
}

\begin{document}

\newcommand{\ra}{\rightarrow}
\newcommand{\Ra}{\Rightarrow}
\newcommand{\UL}{\underline}
\newcommand{\MC}{\multicolumn}
\newcommand{\MR}{\multirow}
\newcommand{\ds}{\displaystyle}
\newcommand{\tabcc}[1]{\begin{tabular}[c]{c} #1 \end{tabular}}
\newcommand{\todo}[1]{\\\textbf{** #1 **}\\}
\newcommand{\rcv}{Reuters Corpus Volume~2}
\newcommand{\mldcc}{Multilingual Document Classification Corpus}

\abstract{
  Cross-lingual document classification aims at training a document classifier on
  resources in one language and transferring it to a different language without
  any additional resources.
  Several approaches have been proposed in the literature and the current best
  practice is to evaluate them on a subset of the \rcv{}. However, this subset covers only few languages (English, German, French and Spanish) and almost all published works focus on the the transfer between English and German.  In addition, we have observed that the class prior distributions differ significantly between the languages. We argue that this complicates the evaluation of the multilinguality.
  \\
  In this paper, we propose a new subset of the Reuters corpus with balanced
  class priors for eight languages. By adding Italian, Russian, Japanese and
  Chinese, we cover languages which are very different with respect to syntax,
  morphology, etc.  We provide strong baselines for all language transfer
  directions using multilingual word and sentence embeddings respectively.
  Our goal is to offer a freely available framework to evaluate cross-lingual
  document classification, and we hope to foster by these means, research in this
  important area.
  \\ \newline
  \Keywords{
    cross-lingual document classification,
    multilinguality,
    \rcv{} (RCV2),
    evaluation framework
  }
}

\maketitleabstract

\section{Introduction}

There are many tasks in natural language processing which require the
classification of sentences or longer paragraphs into a set of predefined
categories.  Typical applications are for instance topic identification
(e.g. sports, news, $\ldots$) or product reviews (positive or negative).
There is a large body of research on approaches for document classification.
An important aspect to compare these different approaches is the availability
of high quality corpora to train and evaluate them.  Unfortunately, most of
these evaluation tasks focus on the English language only, while there is an
ever increasing need to perform document classification in many other
languages.  One could of course collect and label training data for other
languages, but this would be costly and time consuming.
An interesting alternative is \textit{``cross-lingual document
classification''}.  The underlying idea is to use a representation of the words
or whole documents which is independent of the language. By these means, a
classifier trained on one language can be transferred to a different one,
without the need of resources in that transfer language.  Ideally, the
performance obtained by cross-lingual transfer should be as close as possible to
training the entire system on language specific resources.
Such a task was first proposed by \cite{Klementiev:2012:coling_reuters} using
the \rcv{}.  The aim was to first train a classifier on English and then to transfer
it to German, and vice versa.  An extension to the transfer
between English and French and Spanish respectively was proposed by \cite{Mougadala:2016:naacl_biword}.
However, only few comparative results are available for these transfer directions.

The contributions of this work are as follows. We extend previous works and use
the data in the \rcv{} to
define new cross-lingual document classification tasks for eight very different
languages, namely English, French, Spanish, Italian, German, Russian, Chinese and Japanese.
For each language, we define a train, development and test
corpus.
We also provide strong reference results for all transfer directions between
the eight languages, e.g. not limited to the transfer between a foreign language
and English.
We compare two approaches, based either on multilingual word or sentence
embeddings respectively.  By these means, we hope to define a clear evaluation
environment for highly multilingual document classification.

\section{Corpus description}

The \rcv{} \cite{Lewis:Reuters:2004}, in short RCV2\footnote{\url{http://trec.nist.gov/data/reuters/reuters.html}}, is a multilingual corpus with a collection of 487,000 news stories.
Each news story was manually classified into four
hierarchical groups: CCAT (Corporate/Industrial), ECAT (Economics), GCAT (Government/Social) and MCAT (Markets).  Topic codes were assigned to capture the major subject of the news story.
The entire corpus covers thirteen languages, i.e. Dutch, French, German, Chinese, Japanese, Russian, Portuguese, Spanish, Latin American Spanish, Italian, Danish, Norwegian, and Swedish, written by local reporters in each language. The news stories are not parallel. 
Single-label stories, i.e. those labeled with only one topic out of the four top categories, are often used for evaluations. However, the class distributions vary significantly across all the thirteen languages (see Table~\ref{TabEmpPriors}). Therefore, using random samples to extract evaluation corpora may lead to very imbalanced test sets, i.e. undesired and misleading variability among the languages when the main focus is to evaluate cross-lingual transfer.

\begin{table}[t]
  \centering
  \begin{tabular}[t]{|l||*{4}{c|}}
    \hline
    & \MC{4}{c|}{Category} \\
    Language & ECAT & CCAT & GCAT & MCAT \\
    \hline
    \hline
    English & ~6.2\% & 39.8\% & 29.5\% & 24.5\% \\
    German  & ~6.4\% & 30.1\% & 40.9\% & 22.6\% \\
    French  & ~6.3\% & 21.6\% & 60.2\% & 11.8\% \\
    Spanish & ~8.6\% & 15.0\% & ~9.0\% & 67.3\% \\
    Chinese & 19.7\% & 18.2\% & ~2.8\% & 59.4\% \\
    Italian & 18.0\% & 35.7\% & ~9.5\% & 36.8\% \\
    Japanese & 14.8\% & 42.1\% & ~7.0\% & 36.1\% \\
    Russian & 26.9\% & 27.6\% & 13.4\% & 32.2\% \\
    Danish & ~7.5\% & 56.6\% & ~5.3\% & 30.6\% \\
    \hline
  \end{tabular}
  \caption[]{
    Class distribution of all single-label stories per language
    of the entire \rcv{}.
  }
  \label{TabEmpPriors}
\end{table}

\subsection{Cross-lingual document classification}

A subset of the English and German sections of RCV2 was
defined by \cite{Klementiev:2012:coling_reuters} to evaluate cross-lingual
document classification.
This subset was used in several follow-up works and many comparative results are available for the transfer between German and English.
\cite{Mougadala:2016:naacl_biword} extended the use of RCV2 
for cross-lingual document classification to the French and Spanish language
(transfer from and to English).
An analysis of these evaluation corpora has shown that the class prior distributions vary
significantly between the classes (see Table~\ref{TabPriors}). For German and
English, more than 80\% of the examples in the test set belong to the classes
\texttt{GCAT} and \texttt{MCAT} and at most 2\% to the class \texttt{CCAT}.
These class prior distributions are very different for French and Spanish: the
class \texttt{CCAT} is quite frequent with 21\% and 15\% of the French and Spanish
test set respectively.
One may of course argue that variability in the class prior distribution is typical for
real-world problems, but this shifts the focus from a high quality cross-lingual transfer to \textit{``tricks''} for how to best handle the class imbalance.
Indeed, in previous research the transfer between English and German achieves accuracies higher than 90\%, while the performance is below 80\% for EN/FR or even 70\% EN/ES. We have seen experimental evidence that these important differences are likely to be caused by the discrepancy in the class priors of the test sets.

\subsection{Multilingual document classification}

In this work, we propose a new evaluation framework for highly multilingual document classification which significantly extends the current state. We continue to use \rcv{}, but based on the above mentioned limitations of the current subset of RCV2, we propose new tasks for cross-lingual document classification.  The design choices are as follow:

\begin{itemize}
  \item \textbf{Uniform class coverage:}
    we sample from RCV2 the same number of examples for each class and language;
  \item \textbf{Split the data into train, development and test corpus:}
  for each languages, we provide training data of different sizes
	(1k, 2k, 5k and 10k stories),
	a development (1k) and a test corpus (4k);
  \item \textbf{Support more languages:}
    German (DE), English (EN), Spanish (ES), French (FR), Italian (IT), Japanese (JA),  Russian (RU) and Chinese (ZH).
	Reference baseline results are available for all languages.
\end{itemize}

\begin{table}[t!]
  \centering
  \begin{tabular}[t]{|l||*{4}{c|}}
    \hline
    & \MC{4}{c|}{Category} \\
    Language & ECAT & CCAT & GCAT & MCAT \\
    \hline
    \hline
    English & 18.6\% & ~1.5\% & 33.0\% & 46.6\% \\
    German  & 11.9\% & ~0.6\% & 40.6\% & 46.8\% \\
    French  & ~6.0\% & 21.4\% & 60.8\% & 12.8\% \\
    Spanish & ~9.2\% & 14.8\% & ~9.1\% & 66.8\% \\
    \hline
  \end{tabular}
  \caption[]{
    Class distribution of the test set of the RCV2 subsets as used in
    previous publications on cross-lingual document classification.}
  \label{TabPriors}
\end{table}
Most works in the literature use only 1~000 examples to train the document
classifier.  To invest the impact of more training data, we also provide
training corpora of 2~000, 5~000 and 10~000 documents.\footnote{With the
exception of Spanish (9~458 documents) and Russian (5~216 documents) for which
not enough data is available.}
The development corpus for each language is composed of 1~000 and the test set
of 4~000 documents respectively.  All have uniform class distributions.
An important aspect of this work is to provide a framework to study and
evaluate cross-lingual document classification for many language pairs.  In
that spirit, we will name this corpus \textit{``\mldcc{}''},
abbreviated as MLDoc.
The full \rcv{} has a special license and we can not distribute it ourselves.
Instead, we provide tools to extract all the subsets of MLDoc at \url{https://github.com/facebookresearch/MLDoc}.

\section{Baseline results}

\begin{table}[b!]
  \centering
  \begin{tabular}{|l||c|c|c|}
    \hline
    Transfer  & Model & Model & \MR{2}{*}{Evaluation} \\
    type  & training & selection &  \\
    \hline
    \hline
    Zero shot & Train L$_1$ & Dev L$_1$
     & \tabcc{Test L$_i$ } \\
    Targeted   & Train L$_1$ & Dev L$_2$ & Test L$_2$ \\
    Joint      & Train L$_i$ & Dev L$_i$ & Test L$_i$ \\
    \hline
  \end{tabular}
  \caption{Different schemes of cross- and multilingual document classification.}
  \label{TabSchemes}
\end{table}

\begin{table*}[t]
  \centering
  \begin{tabular}[t]{|c||c|*{8}{@{\,\,}c@{\,\,}|}|c|c|}
    \hline
    Training & Develop & \MC{8}{c||}{Accuracy on test languages} & \MC{2}{c|}{Average}\\
    Language & Accuracy & DE$_1$ & EN$_{1,2}$ & ES$_{1,2}$ & FR$_{1,2}$ & IT$_1$ & RU$_2$ & ZH$_2$ & JA & lang$_1$ & lang$_2$ \\
    \hline
    \hline
    \MC{12}{|l|}{\bf MultiCCA word embeddings, aggregation by convolutional network} \\
    German   & 92.2 & (93.7) & 55.95 & \it 73.23 & 71.55 & \it 63.98 & 44.83 & 55.45 & 60.18 & 71.68 & 60.20 \\
    English   & 93.9 & \UL{\bf 81.2} & (92.2) & \UL{\it 72.50} & \UL{72.38} & \UL{\bf 69.38} & \UL{\it 60.80} & \UL{\bf 74.73} & \UL{\bf 67.63} & \UL{\bf 77.52} & \UL{\bf 73.44} \\
    Spanish   & 95.3 & 55.8 & \UL{74.0} & (94.45) & 65.63 & 58.35 & 45.53 & 41.63 & 43.40 & 69.63 & 67.58 \\
    French    & 91.5 & 53.7 & 64.8 & 65.40 & (92.05) & 61.15 & 40.75 & 38.35 & 37.75 & 67.43 & 64.83 \\
    Italian   & 85.6 & 49.2 & 53.7 & 58.68 & 62.25 & (85.55) & 35.58 & 32.13 & 45.30 & 61.87 & 64.83 \\
    Russian   & 86.8 & 40.3 & 72.5 & 41.03 & 44.60 & 42.70 & (85.65) & 42.38 & 39.68 & 48.22 & 57.30 \\
    Chinese   & 90.8 & 48.7 & 56.0 & 35.53 & 53.58 & 47.18 & 40.45 & (87.30) & 50.63 & 48.19 & 46.55 \\
    Japanese  & 87.3 & 52.7 & 54.9 & 54.28 & 48.30 & 44.33 & 40.85 & 44.78 & (85.35) & 50.89 & 48.52 \\
    \hline
    \hline
    \MC{12}{|l|}{\bf Joint sentence embeddings BiLSTM + max pooling, trained on Europarl} \\
    German    & 94.3 & (92.03) & 71.52 & \UL{\bf 75.50} & \UL{\bf 75.45} & 56.45 & - & - & - & 74.15 & - \\
    English   & 90.7 & 71.83 & (88.40) & 66.65 & 72.83 & 60.73 & - & - & - & 72.09 & - \\
    Spanish   & 88.2 & 71.05 & 62.70 & (88.28) & 62.67 & 57.93 & - & - & - & 68.53 & - \\
    French   & 90.6 & \UL{\it 78.42} & \UL{\bf 76.00} & 70.70 & (89.75) & \UL{63.70} & - & - & - & \UL{\it 75.71} & - \\
    Italian   & 83.1 & 66.22 & 67.15 & 67.07 & 65.07 & (82.88) & - & - & - & 69.68 & - \\
    \hline
    \MC{12}{|l|}{\bf Joint sentence embeddings BiLSTM + max pooling, trained on United Nations} \\
    English   & 91.3 & - & (88.83) & 69.50 & \UL{\it 74.52} & - & \UL{\bf 61.42} & \UL{\it 71.97} & - & - & \UL{\bf 73.25} \\
    Spanish   & 86.8 & - & 61.65 & (87.67) & 61.62 & - & 45.10 & 59.88 & - & - & 63.18 \\
    French    & 90.5 & - & \UL{\it 75.35} & \UL{71.80} & (89.55) & - & 59.55 & 69.08 & - & - & 73.07 \\
    Russian   & 83.8 & - & 68.53 & 65.18 & 65.90 & - & (81.60) & 59.65 & - & - & 68.17 \\
    Chinese   & 90.4 & - & 66.30 & 64.78 & 63.82 & - & 54.57 & 87.10 & - & - & 67.31 \\
    \hline
  \end{tabular}
  \caption[]{
    Baseline classification accuracies for \textbf{zero-shot transfer} on the test set of
    the proposed \mldcc{}.
    All classifiers were trained on 1~000 news stories and model selection is performed on
    the Dev corpus of the training language. The same system is then applied to all test languages.
    Underlined scores indicate the best result on each transfer language for each group,
    bold scores the overall best accuracy, and italic ones the second best results.
  }
  \label{TabResZeroShot}
\end{table*}

In this section, we provide comparative results on our new \mldcc{}.  Since the initial work by \cite{Klementiev:2012:coling_reuters} many alternative approaches to cross-lingual document classification have been developed. We will encourage the respective authors to evaluate their systems on MLDoc. We believe that a large variety of transfer language pairs will give valuable insights on the performance of the various approaches.

In this paper, we propose initial strong baselines which represent two complementary directions of research: one based on the aggregation of multilingual word embeddings, and another one, which directly learns multilingual sentence representations.
Details on each approach are given in section~\ref{SectWordRepr} and \ref{SectSentRepr} respectively.
In contrast to previous works on cross-lingual document classification with RVC2, we explore training the classifier on all languages and transfer it to all others, ie. we do not limit our study to the transfer between English and a foreign language.

One can envision several ways to define cross-lingual document classification, in function of the resources which are used in the source and transfer language (see Table~\ref{TabSchemes}). 
The first scheme assumes that we have no resources in the transfer language at all, neither labeled nor unlabeled. We will name this case \textbf{``zero-shot cross-lingual document classification''}. To simplify the presentation, we will assume that we transfer from English to German.
The training and evaluation protocol is as follows.
First, train a classifier using resources in the source language only, eg. the training and development corpus are in English. All meta parameters and model choices are performed using the English development corpus. Once the best performing model is selected, it is applied to the transfer language, eg. the German test set. Since no resources of the transfer language are used, the same system can be applied to many different transfer languages. This type of cross-lingual document classification needs a very strong multilingual representation since no knowledge on the target language was used during the development of the classifier.

In a second class of cross-lingual document classification, we may aim in improving the transfer performance by using a limited amount of resources in the target language. In the framework of the proposed MLDoc we will use the development corpus of target language for model selection. We will name this method \textbf{``targeted cross-lingual document classification''} since the system is tailored to one particular transfer language. It is unlikely that this system will perform well on other languages than the ones used for training or model selection.

If the goal is to build one document classification system for many languages, it may be interesting to use already several languages during training and model selection. To allow a fair comparison, we will assume that these multilingual resources have the same size than the ones used for zero-shot or targeted cross-language document classification, e.g. a training set composed of five languages with 200 examples each.
This type of training is not a cross-lingual approach any more. Consequently, we will refer to this method as \textbf{``joint multilingual document classification''}.

\subsection{Multilingual word representations}
\label{SectWordRepr}

Several works have been proposed to learn multilingual word embeddings, which are then combined to perform cross-lingual document classifications. These word embeddings are trained on either word alignments or sentence-aligned parallel corpora. To provide reproducible benchmark results, we use MultiCCA word embeddings published by \cite{ammar2016massively}. 

There are multiple ways to combine these word embeddings for classification. We train a simple one-layer convolutional neural network (CNN) on top of the word embeddings, which has shown to perform well on text classification tasks regardless of training data size \cite{kim2014convolutional}. Specifically, convolutional filters  are applied to windows of word embeddings, with  a max-over-time pooling on top of them.  
We freeze the multilingual word embeddings while only training the classifier. Hyper-parameters such as convolutional output dimension, window sizes are done by grid search over the Dev set of the same language as the train set. 

\subsection{Multilingual sentence representations}
\label{SectSentRepr}

\begin{table}[t]
  \centering
  \begin{tabular}[t]{|c||*{5}{@{\,}c@{\,}|}|c|}
    \hline
    \MR{2}{*}{Train} & \MC{5}{c||}{Accuracy on test languages} & \MR{2}{*}{Avg} \\
     & DE & EN & ES & FR  & IT &  \\
    \hline
    \hline
    \MC{7}{|l|}{\bf Joint sentence embeddings (Europarl)} \\
    DE & (92.03) & 76.48 & 76.95 & 76.72 & \UL{66.27} & 77.69 \\
    EN & 81.17 & (88.40) & 70.75 & \UL{77.80} & 62.35 & 76.09 \\
    ES & 77.38 & 67.58 & (88.28) & 67.92 & 64.07 & 73.05 \\
    FR & \UL{82.78} & \UL{76.72} & \UL{76.97} & (89.75) & 64.07 & \UL{78.06} \\
    IT & 77.10 & 72.70 & 72.60 & 76.97 & (82.88) & 76.45 \\
    \hline
  \end{tabular}
  \caption[]{
    Baseline classification accuracies for \textbf{targeted transfer} on
    the test set of the proposed MLDoc.
    All classifiers were trained on 1~000 news stories and model selection is
    performed on the Dev corpus of the target language. Each entry corresponds
    to a specifically optimized system.
  }
  \label{TabResTarget}
\end{table}

A second direction of research is to directly learn multilingual sentence
representations.
In this paper, we evaluate a recently proposed technique to learn joint
multilingual sentence representations \cite{Schwenk:2017:repl4nlp}.
The underlying idea is to use multiple
sequence encoders and decoders and to train them with aligned corpora from the
machine translation community. The goal is that all encoders share the same
sentence representation, i.e. we map all languages into one common space.  A
detailed description of this approach can be found in
\cite{Schwenk:2017:repl4nlp}.
We have developed two versions of the system:
one trained on the Europarl
corpus \cite{Koehn:2005:mtsummit_eurparl} to cover the languages
English, German, French, Spanish and Italian, and another one trained on the
United Nations corpus \cite{Ziemski:2016:lrec_unv1} which allows to learn a
joint sentence embedding for English, French, Spanish, Russian and Chinese.
We use a one hidden-layer MLP as classifier.
For comparison, we
have evaluated its performance on the original subset of RCV2 as used in previous publications on
cross-lingual document classification: we are able to outperform the current state-of-the-art in three out of six transfer directions.

\subsection{Zero-shot cross-lingual document classification}

The classification accuracy for \textbf{zero-shot transfer} on the test set of our \mldcc{} are summarized in Table~\ref{TabResZeroShot}.
The classifiers based on the MultiCCA embeddings perform very well on the development corpus (accuracies close or exceeding 90\%). The system trained on English also achieves excellent results when transfered to a different languages, it scores best for three out of seven languages (DE, IT and ZH).\footnote{We exclude Japanese from the comparison since we do not have joint sentence embeddings for that language yet.}
However, the transfer accuracies are quite low when training the classifiers on other languages than English, in particular for Russian, Chinese and Japanese.

The systems using multilingual sentence embeddings seem to be overall more robust and less language specific. They score best for four out of seven languages (EN, ES, FR and RU). Training on German or French actually leads to better transfer performance than training on English.
Cross-lingual transfer between very different languages like Chinese and Russian also achieves remarkable results.

\subsection{Targeted cross-lingual document classification}

\begin{table}[t]
  \centering
  \begin{tabular}[t]{|c||*{5}{@{\,\,}c@{\,\,}|}|c|}
    \hline
    Train & \MC{5}{c||}{Accuracy on test languages} & \MR{2}{*}{Average} \\
    Size & DE & EN & ES & FR & IT &  \\
    \hline
    \hline
    \MC{7}{|l|}{\bf MultiCCA word embeddings} \\
    1k & \UL{91.23} & 79.08 & \UL{86.95} & 81.70 & \UL{77.58} & \UL{83.31} \\
    \hline
    \MC{7}{|l|}{\bf Joint sentence embeddings (Europarl)} \\
    1k & 88.02 & \UL{82.42} & 80.12 & \UL{84.55} & 75.08 & 82.04 \\
    \hline
  \end{tabular}
  \caption[]{
    Baseline classification accuracies on the test set of the proposed
    MLDoc for \textbf{joint multilingual training}.
    Train and test sets are composed of 200 examples form each of the five languages.
  }
  \label{TabResJoint}
\end{table}

The classification accuracy for \textbf{targeted transfer} are summarized in Table~\ref{TabResTarget}. Due to space constraints, we provide only the results for multilingual sentence embeddings and five target languages.
Not surprisingly, targeting the classifier to the transfer language can lead to important improvements, in particular when training on Italian.

\subsection{Joint multilingual document classification}

The classification accuracies for \textbf{joint multilingual training} are given in Table~\ref{TabResJoint}. We use a multilingual train and Dev corpus composed of 200 examples of each of the five languages. One could argue that the data collection and annotation cost for such a corpus would be the same than producing a corpus of the same size in one language only.
This leads to important improvement for all languages, in comparison to zero-shot
or targeted transfer learning.

\section{Conclusion}

We have defined a new evaluation framework for cross-lingual document classification in eight languages. This corpus largely extends previous corpora which were also based on the \rcv{}, but mainly considered the transfer between English and German.
We also provide detailed baseline results using two competitive approaches (multilingual word and sentence embeddings, respectively), for cross-lingual document classification between all eight languages.
This new evaluation framework is freely available at \url{https://github.com/facebookresearch/MLDoc}.

\section{Bibliographical References}
\small
\bibliographystyle{lrec}
\bibliography{refs}

\end{document}